\documentclass{article}
\usepackage{spconf,amsmath,graphicx}
\usepackage[utf8]{inputenc}
\usepackage[english]{babel}
\usepackage{xcolor}
\usepackage{booktabs}
\usepackage{siunitx}
\usepackage{lineno}

\title{Audio-Visual Speech Recognition is Worth 32$\times$32$\times$8 Voxels}

\name{Dmitriy Serdyuk, Otavio Braga, Olivier Siohan}
\address{Google, 111 8th Ave, New York, 10011 USA\\
         \{dserdyuk,obraga,siohan\}@google.com}

\usepackage{amsmath,amsfonts,bm}

\def\eqref#1{equation~\ref{#1}}

\def\1{\bm{1}}

\DeclareMathAlphabet{\mathsfit}{\encodingdefault}{\sfdefault}{m}{sl}
\SetMathAlphabet{\mathsfit}{bold}{\encodingdefault}{\sfdefault}{bx}{n}
\newcommand{\tens}[1]{\bm{\mathsfit{#1}}}
\def\tA{{\tens{A}}}

\def\tF{{\tens{F}}}

\def\tV{{\tens{V}}}

\newcommand{\R}{\mathbb{R}}

\begin{document}
\topmargin=0mm

\maketitle

\begin{abstract}
Audio-visual automatic speech recognition (AV-ASR) introduces the video modality into the speech recognition
process, often by relying on information conveyed by the motion
of the speaker's mouth.
The use of the video signal requires extracting visual features,
which are then combined with the acoustic features to build an AV-ASR
system~\cite{Makino2019-lm}. This is traditionally done with some form of 3D
convolutional network (e.g. VGG) as widely used in the computer vision community. 
Recently,
image transformers~\cite{Dosovitskiy2020-nh} have been introduced to
extract visual features useful for image classification tasks.
In this work, we propose to replace the 3D convolutional visual front-end
with a video transformer
front-end. We train our systems on a large-scale dataset composed of
YouTube videos and evaluate performance on the publicly available
LRS3-TED set, as well as on a large set of YouTube videos. On a
lip-reading task, the transformer-based front-end shows superior
performance compared to a strong convolutional baseline. 
On an AV-ASR task, the transformer front-end performs as well as
(or better than) the convolutional baseline.
Fine-tuning our model on the LRS3-TED training set matches previous
state of the art.
Thus, we experimentally show the viability of the convolution-free model for AV-ASR.
\end{abstract}


\section{Introduction}
\label{sec:intro}

In many automatic speech recognition (ASR) applications, the audio signal originates 
from an audio-visual source (e.g. YouTube video, TV broadcast). 
The visual signal
provides both contextual information that may be related to the spoken content
(e.g. what is being discussed may be visually present on the
video) as well as information conveyed by the motion
of the speaker's mouth. 
This can be integrated into conventional ASR, leading to \emph{audio-visual automatic speech recognition}~(AV-ASR)~\cite{Neti2000-ca,Gupta2017-lz,Makino2019-lm}.

In this work, we exploit the information provided by the
visible motion of the speaker's speech production articulators, using video crops centered around the speaker's mouth.
When this visual information alone is being used for recognition, the task
is known as \emph{lip reading}~\cite{Afouras2019-jp}.
Because the visual information is not affected by ambient noise, it has been
shown~\cite{Makino2019-lm} to improve the robustness of ASR systems under adverse conditions.

Lip-reading and AV-ASR require extracting visual features from the video stream.
This is often done
using a trainable 3D convolutional network operating both in time and spatially
over the image axes, similar to VGG features used in computer
vision~\cite{Simonyan2015-tq}, and has led to the design of state-of-the-art AV-ASR and lip-reading systems~\cite{Makino2019-lm,Afouras2018-gl}.

Recently, a series of breakthroughs were made 
due to the development of the attention-based transformer architecture~\cite{Vaswani2017-di}.
It has been shown that transformer models are beneficial
to a variety of sequence-to-sequence learning tasks such as NLP~\cite{Devlin2018-gz, Radford2019-wr} and
speech recognition~\cite{Zhang2020-nr}.
This was adopted in computer vision as well, as illustrated by the work on
\emph{vision transformers} (ViT,~\cite{Dosovitskiy2020-nh}).
In a nutshell, the ViT model extracts non-overlapping 16x16 image patches,
embeds them with a linear transform, and runs a transformer.
That work emphasises that the ViT model has fewer inductive biases
than convolutional networks.
Therefore, given enough training data, ViT is able to learn features that
improve the performance on a variety of image processing tasks.
The ViT architecture was extended to video inputs in~\cite{Arnab2021-mq, Bertasius2021-ke}.
Hence, a number of hard vision tasks that were thought to require convolutions
were tackled using transformers.

In this work, we focus on the vision task of embedding the visual features for AV-ASR.
Our paper aims to test and explore the viability of using a  \emph{fully transformer-based} architecture, where both the video and audio front-ends are transformer networks.
We believe that such architectures are important to research because
they have fewer inductive biases and should be able to better fit the training data.
Therefore, we propose to use a transformer video encoder to learn visual features for AV-ASR and lip reading, keeping the rest of the model architecture unchanged.
We compare the proposed approach with our baseline system that uses a VGG convolutional network to extract visual features, similar to the work done in~\cite{Makino2019-lm}.
We design a video transformer akin to~\cite{Arnab2021-mq}, but tailored specifically for an AV-ASR task.
We extract \emph{tubelets} (voxels, or in other words, 3D~patches) from the video input and embed each of them with a shared linear transform.
The transformer is applied to the embedded patches combined with the positional information.

\newpage
The contributions of this paper are:
\begin{itemize}
    \item The design and use of a fully transformer-based architecture for visual feature extraction for lip reading and A/V ASR applications. The design choices for our model, as well as
    a strong convolutional baseline are reported in the Section~\ref{sec:model};
    \item The evaluation of our models against a competitive baseline and prior works (Section~\ref{sec:exp}). 
    In particular, we use a lip-reading task to assess the model's ability to encode video
    (Section~\ref{ssec:exp:lip_reading}), and report 
    up to 8\% relative improvement over a convolutional baseline.
    In addition, we evaluate the proposed model on a real-world AV-ASR task (Section~\ref{ssec:exp:av_asr}), 
    where the transformer shows at least a similar result as the convolution.
\end{itemize}

\section{Related Work}
\label{sec:related}

\paragraph*{Audio-Visual Automatic Speech Recognition.}
Audio-visual speech recognition~\cite{Neti2000-ca} made significant
progress thanks to the introduction of the end-to-end approaches~\cite{Assael2016-zg, Chung2016-bd, Makino2019-lm}.
Using deep neural networks and end-to-end training allowed these and other works
to tackle the audio-visual speech recognition ``in the wild'',
i.e. unconstrained open-world utterances.

Recently, an outstanding work~\cite{Ma2021-al} achieved the state of the art on the tasks
of audio-visual speech recognition and lip reading.
The work used a combination of the CTC~\cite{Graves2006-vf} and seq2seq~\cite{Bahdanau2014-vc}
losses with a conformer network~\cite{Gulati2020-jh}.
Our work matches this state of the art on the LRS3-TED eval set.

\paragraph*{Transformer-based Models for Video.}
Since the attention-based~\cite{Bahdanau2014-vc} transformer architecture was introduced in~\cite{Vaswani2017-di},
it quickly became a model of choice for natural language processing~\cite{Devlin2018-gz, Radford2019-wr}.
Later, it was employed for other sequential tasks, such as speech recognition~\cite{Zhang2020-nr}.
A highly influential paper~(ViT,~\cite{Dosovitskiy2020-nh}) was the first work demonstrating that
the transformer architecture performs at least as well as convolutions.

The visual transformer was extended to the video~\cite{Arnab2021-mq, Sharir2021-mh, Neimark2021-md}, in particular for the tasks such as video classification and action recognition.
In contrast to these works, this paper focuses on a sequence-to-sequence task.
While the sequence output provides a signal stronger than classification,
the task is harder due to the fact that the network is required to learn
the alignment.

\section{Model}
\label{sec:model}

\begin{figure}
    \centering
    \includegraphics[width=0.3\textwidth]{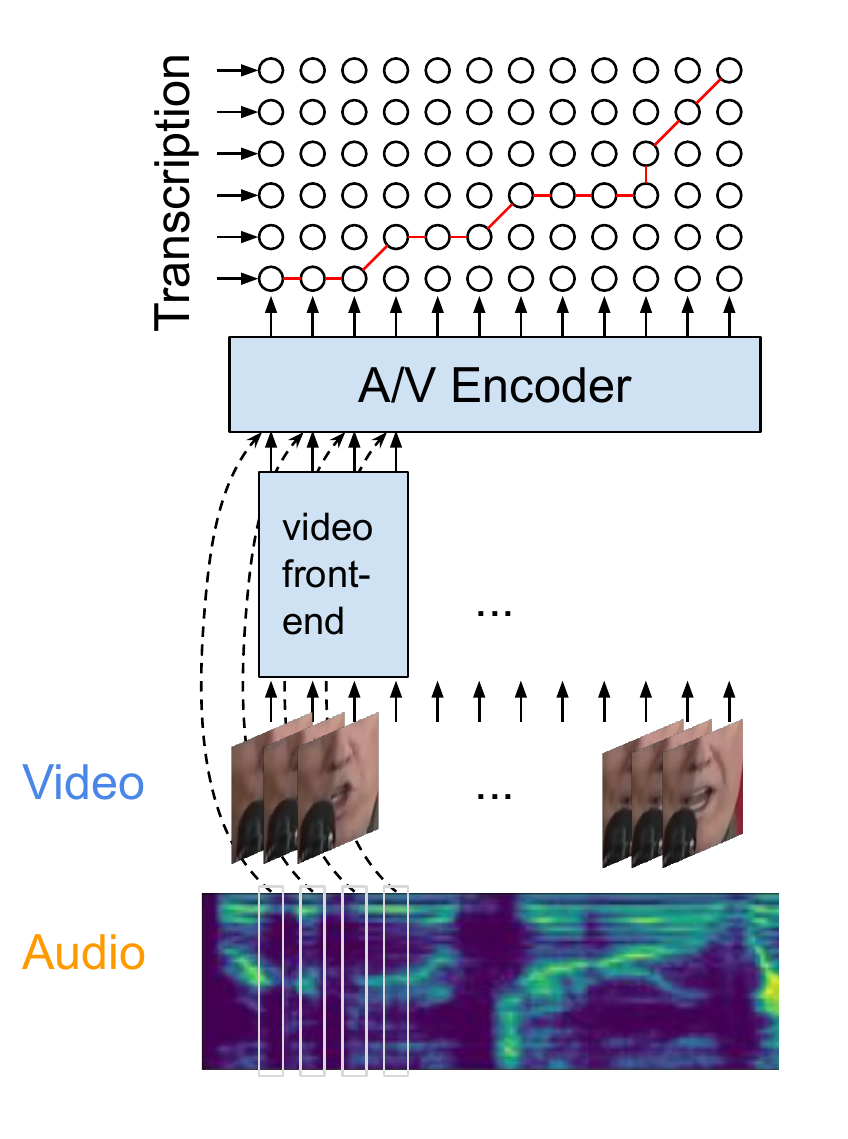}
    \caption{An overview of end-to-end AV-ASR and lip reading models.
             The video is encoded with a video front-end.
             The visual features and the acoustic features are concatenated
             and fed through the AV encoder to be used for the RNN-T loss.}
    \label{fig:avasr}
\end{figure}

In this section, we present an overview of the shared ASR model architecture used in all of our experiments, as well as the two video front-ends we are comparing in the present work: a state-of-the art baseline based on purely (2+1)D convolutional blocks~\cite{Tran2017-nn}, and our proposed front-end, which employs self-attention on video patches through transformer blocks instead of convolutions.

\subsection{Common A/V ASR Model Architecture}

Besides the video front-end, we share the same model architecture illustrated in Figure~\ref{fig:avasr} in all our experiments, and we outline the main components next.

\

\noindent
{\bf Acoustic Features.} We employ mel filterbank features as acoustic features. The 16kHz-sampled input audio is framed with 25ms windows smoothed with the Hann window function, with steps of 10ms between consecutive frames. We compute energies in 80 mel filter bank channels at each frame, compressing their range with a $\log$ function. We then fold every 3 consecutive feature vectors together, yielding a 240 dimensional feature vector every 30ms, which corresponds to acoustic features at about 33.3Hz. We denote the input acoustic features tensor by $\tA \in \R^{B\times T \times D_A}$, where $B$ is the batch size, $T$ is the number of time steps and $D_A$ ($=240$) the dimension of the acoustic features.

\

\noindent
{\bf Visual Features.} The videos in our training set have frame rates ranging from around 23 to 30 fps, so in order to make the input uniform we synchronize the videos with the acoustic features by resampling the video with nearest neighbor interpolation in time at the acoustic features sample rate (33.3Hz). In the spatial dimension, we crop the full face tracks around the mouth region to generate images of resolution $128\times128$, with RGB channels normalized between $-1$ and $1$.

We then extract visual features from the synchronized mouth track video using a \emph{video front-end}, which outputs a tensor $\tV \in \R^{M\times T\times D_v}$. An example of a video front-end is a 3D ConvNet \cite{LeCun1998-pe}.

\

\noindent
{\bf Modality fusion. } We combine the visual and acoustic features, which have the same temporal resolution, with simple concatenation. The output of the combined audio-visual frontend is thus a tensor $\tF = [\tA; \tV] \in \R^{M\times T\times (D_a + D_v)}$.

\

\noindent
{\bf Encoder.} The resulting merged audio-visual features $\tF$ are then fed into the \emph{audio-visual encoder}, which consists of a standard $14$-layer Transformer encoder~\cite{Vaswani2017-di}.

\

\noindent
{\bf Decoder.} The output of the AV encoder is fed into an RNN-T~\cite{Graves2006-vf, Graves2012-ad} decoder, consisting of 2 LSTM layers of 2048 units with character tokens.


\subsection{Video Front-Ends}

\subsubsection{(2+1)D ConvNet Baseline}
\label{ssec:model:baseline}

We use a convolutional video front-end baseline for all our experiments.
We build upon a prior work~\cite{Makino2019-lm}.
Our baseline has several modifications.
\cite{Makino2019-lm} uses a 3D~convnet to capture both spatial and temporal features of the video.
We improve the efficiency by decomposing the 3D~convolution into separate spatial and temporal kernels~\cite{Tran2017-nn,Xie2017-vb} (i.e. a kernel of the dimension [3,~3,~3] becomes a kernel of dimension [1,~3,~3] followed by another kernel of dimension [3,~1,~1]).
Thus, we construct a 5-layer VGG-like network and decompose it into a 10-layer\footnote{
Kernel sizes are: 23, 64, 230, 128, 460, 256, 921, 512, 460, 512.} VGG~(2+1)D net.
We apply max pooling of size 2 after each pair of layers (with an exception for layer 4).

\subsubsection{Video Transformer Front-end}
\label{ssec:model:vit}

\begin{figure*}
    \centering
    \includegraphics[width=\textwidth, trim=0 350 0 0,clip]{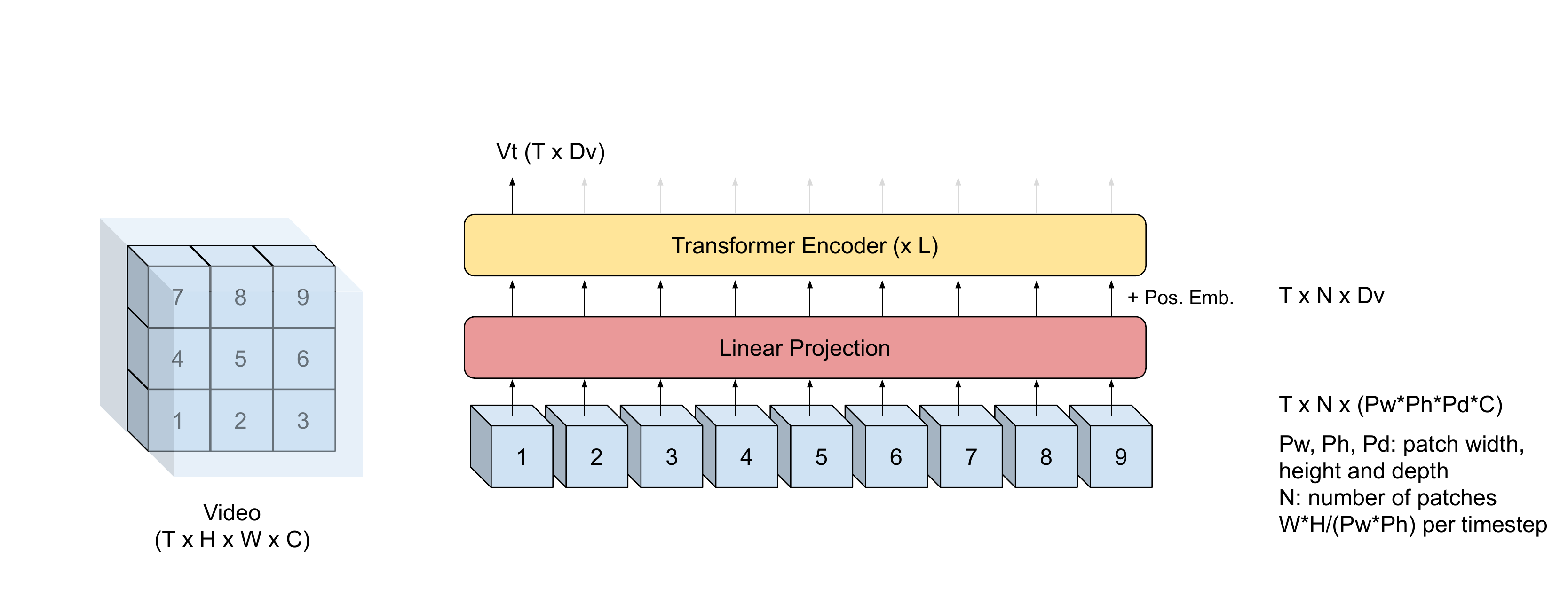}
    \caption{An overview of the proposed architecture for the video-encoding transformer.
             The input video is split into `tubelets'.
             The tubelets are embedded with a linear projection and fed into a transformer.}
    \label{fig:avasr_vit}
\end{figure*}

The proposed architecture is inspired by~\cite{Dosovitskiy2020-nh} and~\cite{Arnab2021-mq}.
We construct a transformer for the feature extraction from the mouth tracks.
The architecture considered in this work is depicted in the Figure~\ref{fig:avasr_vit}.
First, we extract 4-dimensional `tubelets' from the video input (a 3D version of the image patches).
The tubelets are flattened and fed through an affine projection.
These embeddings for each tubelet are combined with a positional embedding~\cite{Shaw2018-px} and fed
into an off-the-shelf transformer.
We use a 6-layer version of the transformer commonly used for the sequence tasks.
This transformer has 8-headed attention and 512-dimensional features.
Finally, we take the first output of the last layer of the transformer and forward it into
the rest of the AV-ASR or lip reading network.

We extract $32\times32\times8$ (H$\times$W$\times$T) tubelets at each time-step.
This corresponds to the temporal stride of 1.
The tubelets do not intersect in the spatial dimensions.
In total, we use 16 (4x4) tubelets at each time-step.
This set of tubelets is fed into the transformer.

\section{Experiments}
\label{sec:exp}

In this section, we first outline our training procedure, with a brief description of our datasets, followed by an evaluation of the proposed transformer visual frontend on two scenarios: First, we evaluate our model on the lip reading scenario, where only the visual signal is present. Then, we evaluate on the audio-visual scenario, where we use both the audio and video signals for speech recognition.

\

\noindent
{\bf Datasets.} In all our experiments we use a training dataset derived from around 90k hours of transcribed, public YouTube videos. We train on short video segments, limited to 512 frames (around 15 seconds), using the semi-supervised procedure originally proposed in \cite{Liao2013-em} and extended in \cite{Makino2019-lm,Shillingford2019-sc} to include video. We extract short segments where the force-aligned user uploaded transcription matches the transcriptions from a production quality ASR system. From these segments we then keep the ones in which the face tracks match the audio with high confidence. Please consult the original references for more details.

In order to obtain the development and the test sets we use a separate set of YouTube videos
which was transcribed by professionals, the YTDEV18 set.
In addition to the YTDEV18 set, we evaluate on the LSR3-TED corpus~\cite{Afouras2018-pq}.
We run the same pipeline for extracting the mouth tracks and the acoustic features.
We found that the LSR3-TED corpus is significantly easier than YTDEV18.
Furthermore, the results on LSR3-TED tend to have high variance.
Despite these downsides, we use this public evaluation set to be able to compare
to prior publications.

\

\noindent
{\bf Training.} For all our models we use the following training schedule.
First, we increase the learning rate linearly up to $1e^{-4}$ for the first 30,000 iterations
and maintain it constant for the next 170,000 iterations.
Then, the learning rate is annealed exponentially down to $1e^{-6}$ for the next 100,000 iterations.
We use the batch size of 1024 and the Adam~\cite{Kingma2014-dh} training algorithm.

\subsection{Lip Reading}
\label{ssec:exp:lip_reading}

The lip reading models are trained on YouTube videos with audio omitted.

The results for the lip reading experiments are summarized in the Table~\ref{tab:lip_reading}.
We observe that the transformer based model ViT 3D outperforms our convolutional baseline.
Furthermore, our models outperform the prior published models~\cite{Afouras2018-gl, Ma2021-al}, 
although we should emphasize that our training data is different from these publications.

The ViT model shows 4\% relative improvement over
the VGG~(2+1)D baseline on the YTDEV18 set and 8\% on the LRS3-TED.
There is 23\% relative improvement comparing to~\cite{Makino2019-lm}
and 40\% comparing to~\cite{Ma2021-al}.
Therefore, we conclude that the video transformer is able to extract rich features from the mouth tracks
and these features can be leveraged by the decoder.

\subsection{Audio-Visual Automatic Speech Recognition}
\label{ssec:exp:av_asr}

Inspired by the positive results for the lip reading, we continue to the AV-ASR task.

\paragraph*{First, we evaluate our AV-ASR models on the YTDEV18 set.}
The results are summarized in the Table~\ref{tab:av_asr}.
The audio is a much stronger signal for the recognition model.
Therefore, despite the fact that the ViT 3D is able to extract better visual features,
it performs just as well as the VGG baseline when the audio is clean.

In order to demonstrate the strength of the transformer video front-end,
we introduce noise into the eval data.
We artificially apply additive noise of magnitudes 20, 10, 0dB.
Finally, we introduce noise by overlapping a random short utterance ($<$5s)
at the start or end of the evaluated utterance.

We observe that the transformer model matches the performance of
the VGG baseline with low amounts of noise.
The ViT model outperforms the baseline in the case of high level of noise (0dB).
Finally, in the case of the overlapped audio, the ViT performance degrades.
We hypothesise that this is due to the more drastic domain shift between
the train data and the test data.
More specifically, the train data is augmented with MTR~\cite{Cui2015-ox},
therefore it is easy to generalize to the additive noise.

The first row of the table reports the performance of an audio-only model.
By removing the video input and the video front-end we demonstrate
the importance of the visual information.
Especially the visual information helps for highly noisy data (0dB set).

\paragraph*{Second, we test our model on a public dataset LRS3-TED.} 
This dataset consists of recordings of TED talks.
The LRS3-TED eval set is smaller than YTDEV18, 
which usually leads to a higher variance in the results.
On the other hand, this eval set is considerably simpler than YTDEV18.
One of the reasons is that the audio quality is high and the video is clean,
high definition, and almost always centered at the speaker.
We use the ``0.4'' version of this dataset,
where the sets of speakers in the train and eval sets are disjoint
and we additionally ensure that the TED videos are excluded from our train set.

We report the results on the LRS3-TED in the Table~\ref{tab:ted_results}.
The proposed model outperforms the convolutional baseline.
However, due to the differences in the training and testing conditions, 
the performance of our vanilla models (VGG~(2+1)D and AV~ViT) is inferior to 
the previous results~\cite{Ma2021-al}.
Therefore, we fine-tune our best model on the 50-50 mix of
our training data and the LRS3-TED train set.
More precisely, we fine-tune the model for 10,000 steps with the mini-batch size of 4096.
We use the learning rate of $1e^{-5}$ and anneal it exponentially down to~$5e^{-8}$.
This fine-tuning process leads to a result comparable to~\cite{Ma2021-al} matching
the state of the art.

\subsection{Computational Performance Analysis}
\label{ssec:exp:performance}

We run a series of benchmarks to measure the computational performance
of the proposed model compared to the baseline.
In order to assess the speed of training, we run the forward propagation
20 times on a mini-batch and report the average metric.
We use the standard TensorFlow utilities to measure the number of floating point operations
and the wall clock time to measure the latency on a TPU.

The results are summarized in the Table~\ref{tab:performance}.
Despite the fact that the transformer front-end requires significantly more
floating point operations, the decrease in latency is small.
This can be attributed to the fact that many transformer computations
can be performed in parallel.
Another important point is that we can fit a much higher number of trainable parameters
into the same memory due to the large feature maps when using convolution.

\begin{table}
    \centering
    \caption{Lip-reading performance, \%WER.}
    \begin{tabular}{lS[table-format=2.2]S[table-format=2.2]}
        \toprule
        \bfseries Model      & {\bfseries YTDEV18} & {\bfseries LRS3-TED} \\ 
        \midrule
        TM-seq2seq~\cite{Afouras2018-gl} & {--} & 58.9 \\
        ResNet+Conf~\cite{Ma2021-al}     & {--} & 43.3 \\
        RNN-T~\cite{Makino2019-lm}       & 48.5 & 33.6 \\
        \midrule
        VGG (2+1)D & 40.5 & 28.2 \\ 
        ViT 3D     & 38.8 & 25.9  \\ 
        \bottomrule
    \end{tabular}
    \label{tab:lip_reading}
\end{table}

\begin{table}
    \centering
    \caption{Audio-visual ASR performance, \%WER. ($\infty$dB) is the clean subset;
             20db, 10dB, 0dB -- data with artificial noise added;
             ``Overlap'' -- contains overlapped utterances.
             }
    \begin{tabular}{lccccc}
        \toprule
        \bfseries Model      & $\infty$dB & 20dB & 10dB & 0dB & Overlap \\
        \midrule
        Audio-only & 16.5 & 17.0 & 19.8 & 42.9 & 35.0 \\
        \midrule
        VGG    & 14.4 & 14.5 & 15.6 & 23.4 & \bfseries 31.2 \\
        AV ViT & 14.4 & 14.6 & 15.6 & \bfseries 23.1 & 31.9 \\
        \bottomrule
    \end{tabular}
    \label{tab:av_asr}
\end{table}

\begin{table}
    \centering
    \caption{AV-ASR performance on the LRS3-TED dataset.
             Models denoted with $^*$ are trained on a large dataset of YouTube videos.
             The last row corresponds to a fine-tuned AV ViT model.}
    \label{tab:ted_results}
    \begin{tabular}{lS[table-format=2.1]}
         \toprule
         \bfseries Model & {WER, \%}  \\
         \midrule
         TM-CTC~\cite{Afouras2018-gl} & 27.7 \\
         EG-s2s~\cite{Xu2020-sf} & 6.8 \\
         RNN-T~\cite{Makino2019-lm}$^*$ & 4.5 \\
         ResNet+Conf~\cite{Ma2021-al} & 2.3 \\
         \midrule
         VGG (2+1)D$^*$ & 3.9 \\
         AV ViT$^*$ & 3.6 \\
         \hspace{0.3cm} + fine-tune & 2.3 \\
         \bottomrule
    \end{tabular}
\end{table}

\begin{table}
    \centering
    \caption{Comparison of the computational performance.
             We measure the speed in terms of the number of
             floating point operations, and in terms of wall clock time.
             We benchmark 20 times and report the average.
             Additionally we report the number of parameters.
             Notice, that we are able to fit significantly more parameters for a transformer model.}
    \label{tab:performance}
    \begin{tabular}{lcS[table-format=2.1]S[table-format=3.1]}
         \toprule
         \bfseries Video front-end & GFLOPS & {Params, M} & {Latency, ms} \\
         \midrule
         VGG (2+1)D & 299.3 &  7.0 & 120.7 \\
         ViT        & 520.7 & 37.2 & 162.3\\
         \bottomrule
    \end{tabular}
\end{table}

\section{Conclusions}
\label{sec:conclusions}

In this work we designed a transformer visual front-end for the AV-ASR task. To the best of our knowledge, the use of a purely transformer-based visual front-end in combination with a transformer encoder makes this the first fully-transformer end-to-end architecture for AV-ASR.
Our proposed model outperforms a strong baseline employing convolutions for the visual frontend on the lip-reading scenario, and matches its performance on the audio-visual ASR scenario.

Finally, we fine-tuned our model on the publicly available LRS3-TED dataset and were able to achieve a new state-of-the-art word error rate on the lip-reading scenario, while matching the performance of the best published model when both acoustic and visual signals are employed.

As a side note, unfortunately we were not able to evaluate on the test sets with BBC data (LRS2 and LRW), datasets which are widely reported in the literature but whose licence prohibits its use by individual scientists and research in industry.

The future work includes using a conformer front-end to further improve the performance.
Other possible directions include extending the architecture to online recognition
and multiple speakers.
For the online recognition, the video transformer would need to take into account
only the past frames.
Tackling multiple speakers would require employing an architecture that can handle  multiple video inputs, such as~\cite{Braga2020-yw,Braga2021-lj}.

\section{Safe AI Principles}
\label{sec:safe_ai}

We are fully aware of the sensitive nature of the audio-visual speech recognition research
and other AI technologies used in this work.
Therefore, we ensure that this work abides by the Google AI Principles~\cite{noauthor_undated-lg}.

\bibliographystyle{IEEEbib}
\bibliography{lit}

\end{document}